\documentclass[11pt,letterpaper]{article}
\usepackage[draft]{hyperref}
\usepackage{acl2012}
\usepackage{booktabs}

\usepackage[printwatermark]{xwatermark}

\usepackage{times}
\usepackage{latexsym}
\usepackage{amsmath}
\usepackage{amsfonts}
\usepackage{amsthm}
\usepackage{subcaption}
\usepackage{tikz}
\usetikzlibrary{bayesnet}
\usepackage{tcolorbox}
\usepackage{adjustbox}
\usepackage{xcolor,lipsum}
\usepackage{adjustbox}
\usepackage{algorithm}
\usepackage[noend]{algpseudocode}
\usepackage{float}
\usepackage{appendix}
\usepackage{url}
\usepackage{microtype}
\usepackage{xfrac}
\usepackage{soul}
\usepackage[safe]{tipa}

\newcommand{\word}[1]{\textit{#1}}
\newcommand{\qtheta}{q_{{\boldsymbol \theta}}}
\newcommand{\vtheta}{{\boldsymbol \theta}}
\newcommand{\pI}{p}

\usepackage[noabbrev,capitalize]{cleveref}

\crefname{equation}{equation}{equations}
\crefname{section}{section}{sections}
\crefname{section}{\S}{\S\S}
\Crefname{section}{\S}{\S\S}
\crefformat{section}{#2\S#1#3}

\usepackage[disable]{todonotes}

\newcommand{\note}[4][]{\todo[author=#2,color=#3,size=\scriptsize,fancyline,caption={},#1]{#4}}
\newcommand{\jason}[2][]{\note[#1]{jason}{green!40}{#2}}
\newcommand{\ryan}[2][]{\note[#1]{ryan}{purple!40}{#2}}
\newcommand{\mans}[2][]{\note[#1]{mans}{orange!40}{#2}}

\newcommand{\Mans}[2][]{\mans[inline,#1]{#2}\noindent}
\newcommand{\Jason}[2][]{\jason[inline,#1]{#2}\noindent}

\newcommand{\vect}[1]{\boldsymbol}

\newcommand*{\mybox}[1]{%
  \fcolorbox{white}{white}{\raisebox{0pt}[0.5\baselineskip][0.05\baselineskip]{%
      \hbox to 1.8cm{\hss#1\hss}}}}

\newcommand*{\myboxgrey}[1]{%
  \fcolorbox{gray!25}{gray!25}{\raisebox{0pt}[0.5\baselineskip][0.05\baselineskip]{%
    \hbox to 1.8cm{\hss#1\hss}}}}

\newwatermark[allpages,color=gray!25,angle=45,scale=3,xpos=0,ypos=0]{PREPRINT}

\title{On the Complexity and Typology of Inflectional Morphological Systems}
\author{{\bf Ryan Cotterell}{\normalfont \raise1.0ex\hbox{\normalsize \textschwa}} \and {\bf Christo Kirov}{\normalfont \raise1.0ex\hbox{\normalsize \textschwa}} \and {\normalfont {\bf Mans Hulden}\raise1.0ex\hbox{\normalsize \textipa{H}}} \and {\normalfont {\bf Jason Eisner}\raise1.0ex\hbox{\normalsize \textschwa}} \\ \raise1.0ex\hbox{\normalsize \textschwa}Department of Computer Science,  Johns Hopkins University, Baltimore MD, 21218 \\
  \raise1.0ex\hbox{\normalsize \textipa{H}}Department of Linguistics,  University of Colorado, Boulder CO, 80309 \\
  \texttt{\{ryan.cotterell,eisner\}@jhu.edu}}

\begin{document}

\maketitle

\begin{abstract}
  We quantify the linguistic complexity of different languages'
  morphological systems.  We verify that there is an empirical
  trade-off between paradigm size and irregularity: a language's
  inflectional paradigms may be either large in size or highly irregular, but never
  both. Our methodology
  measures paradigm irregularity as the entropy of the surface
  realization of a paradigm---how hard it is to jointly predict all
  the surface forms of a paradigm.  We estimate this by a variational
  approximation.  Our measurements are taken on large morphological
  paradigms from 31 typologically diverse languages.
\end{abstract}

\Jason{Would like to see 2 points per language on the graph---nouns and verbs in different colors.  If verb systems seem to have a more generous complexity bound, that might indicate a trade-off between complexity and functionalism.  Also, I'd like to see a paired design: within a language, does increasing one kind of complexity from nouns to verbs (if it was already high on nouns) force us to decrease the other kind?}
\Jason{Note that we are measuring informational complexity, $H(p_{\text{true}})$, estimated as $H(p_{\text{true}},q)$ using the best model $q$ that we can find.  Ideally we would have used cognitive complexity $H(q_{\text{human}})$, based on the model that humans actually find, which might be either worse or better than our current best ML efforts.  Determining humans' model would require nonce word experiments.}

\section{Introduction}
What makes an inflectional system ``complex''?
Linguists have sometimes considered measuring this by the size of the inflectional paradigms---the number of
morpho-syntactic distinctions the language makes \cite{mcwhorter2001world}.
However, this gives only a partial picture of complexity
\cite{sagot2013comparing}; beyond simply being larger, some
inflectional systems are more irregular---it is harder to guess
forms in the paradigm from other forms in the same paradigm. \newcite{ackerman2013} hypothesize that
these two notions of morphological complexity
interact: while a system may be complex along either axis, it is never complex along both, providing a trade-off.
\jason{In short: ``If a system has a lot of forms, they must be
  relatively predictable'' (otherwise too hard to learn).  But we
  wouldn't expect ``a lot of forms'' to be about entropy, but maybe
  about number of forms (at least forms with non-negligible
  probability)?  Try both.  And try a finer-grained prediction: We
  should expect that rare slots should be easy to predict.  A
  hard-to-predict form that becomes rare should get regularized (or
  should become even rarer, leaving a gap).}

In this work, we
develop machine learning tools to operationalize this hypothesis using recurrent neural networks and latent
variable models to measure the complexity of inflectional systems. We
explain our approach to quantifying two aspects of inflectional
complexity and, in one case, derive a variational bound to enable
efficient approximation to the metric. This allows a completely
data-driven approach by which we can measure the morphological
complexity of a given language in a clean, relatively theory-agnostic manner.\looseness=-1

Our study focuses on an evaluation of 31 diverse languages, using collections of orthographic paradigms. Importantly, our method does not require a linguistic analysis of words into their constituent morphemes, e.g., \textit{hoping} $\mapsto$ \textit{hope}$+$\textit{ing}.  We
find support for the hypothesis of \newcite{ackerman2013}.
Concretely, we show that the more forms an inflectional paradigm has,
the more predictable the forms must be from one another (for
example, they might be related by a simple change of suffix).  This
intuition has a long history in the linguistics community, as field
linguists have often noted that languages with extreme morphological
richness, e.g., agglutinative and polysynthetic languages, have
virtually no exceptions or irregular forms. Our contribution lies in
mathematically formulating this notion of regularity and providing
a means to estimate it by fitting a probability model.  Using these
tools, we provide a quantitative verification of this conjecture
on a large set of typologically diverse languages, which is significant
with $p < 0.05$.\ryan{Add actual value!}

\ryan{This isn't discussed yet: Syntagmatic vs paradigmatic morphology. We don't believe there is a
clear distinction here.}

\ryan{This isn't discussed yet: A combinatorial system of morphemes does preclude paradigmatic
organization, and in fact subsumes it in most cases.}

\ryan{Note sure what this means: People can generate not only forms that they've never seen, but
\emph{paradigm cells} that they've never seen. A combinatorial system
over morphemes makes this kind of generalization possible.}

\ryan{Need to address: A major difficulty with the Malouf et al. approach is that it relies
on a linguist to accurately catalog and classify the number of
declensions and exponents used in a morphological system. Any entropy
calculations are strongly affected by this choice of analysis.}

\ryan{Need to mention: We sidestep the issue by letting the different declensions and
inflection patterns remain implicit in the data, and using an RNN to
find find the relevant patterns. This produces an unbiased analysis.}

\section{Morphological Complexity}\label{sec:conjecture}

\subsection{Word-Based Morphology}\label{sec:word-based-morphology}
We adopt the framework of word-based morphology
\cite{aronoff1976word,spencer1991morphological}.\footnote{See \newcite[Part
  II]{baerman2015oxford} for a tour of alternative views of inflectional
  paradigms.}
Thus, for the rest
of the work we will define an \textbf{inflected lexicon} as a set of
word types.
\jason{should recast to say we're using plats, since we basically
  assume this once we get to the model!}
Each word type is a triple of
\begin{itemize}
  \setlength\itemsep{0.5em}
\item a \textbf{lexeme} (an arbitrary integer or string that indexes the
  word's core meaning and part of speech)
\item a \textbf{slot} (an arbitrary integer or object that indicates how the
  word is inflected)
\item a \textbf{surface form} (a string over a fixed phonological or orthographic
  alphabet $\Sigma$)
\end{itemize}

We write $\pi(\ell)$ for the set of word types (triples) in the
lexicon that share lexeme $\ell$, known as the {\bf paradigm} of
$\ell$.  The slots that appear in this set are said to be {\bf filled}
by the corresponding surface forms.  \ryan{More to say about word and
  paradigm morphology.}  For example, in the English paradigm
$\pi(\word{walk}_{\text{Verb}})$, the past-tense slot is filled by
\word{walked}.

Nothing in our method requires a Bloomfieldian structuralist analysis
that decomposes each word into underlying morphemes: rather, this
paper is a-morphous in the sense of
\newcite{anderson1992morphous}.

More specifically, we will work within the UniMorph annotation scheme
\cite{sylak2016composition}.  In the simplest case, each slot
specifies a morpho-syntactic \textbf{bundle} of inflectional features
such as tense, mood, person, number, and gender.  For example, the
Spanish surface form \word{pongas} appears with a slot that indicates
that this word has the features $\left[\right.${\sc tense}$=${\sc
  present}, {\sc mood}$=${\sc subjunctive}, {\sc person}$=${\sc 2},
{\sc number}$=${\sc sg}$\left.\right]$. \jason{this doesn't look like the bundles used as examples later}
However, in a language where two or more feature bundles
systematically yield the same form across all lexemes, UniMorph
generally collapses them into a single slot that realizes multiple
feature bundles.  Thus, a single ``verb lemma'' slot suffices to
describe all English surface forms in \{\word{see}, \word{go},
\word{jump}, \ldots\}: this slot indicates that the word can be a bare
infinitive verb, but also that it can be a present-tense verb that may
have any gender and any person/number pair other than
3rd-person/singular.
We postpone a discussion of the details of
UniMorph until \cref{sec:unimorph}, but it is
mostly compatible with other, similar schemes.

\subsection{Defining Complexity}\label{sec:complexity}
\newcite{ackerman2013} distinguish two types of morphological
complexity, which we elaborate on below. For a more general overview of
morphological complexity, see \newcite{baerman2015understanding}.

\subsubsection{Enumerative Complexity}
The first type, {\bf enumerative complexity} ({\bf e-complexity}), is
the number of morpho-syntactic distinctions a language makes within a
part of speech.  For example, the enumerative complexity of English
verbs can be quantified as the average size of the paradigm
$|\pi(\ell)|$ where $\ell$ ranges over a list of English verb lexemes.

The notion of e-complexity has a long history in linguistics.
The idea was explicitly discussed as early as \newcite{edward1921language}.
More recently, \newcite{sagot2013comparing} has referred to this
concept as {\bf counting complexity}, referencing
 comparison of the complexity of creoles and non-creoles by \newcite{mcwhorter2001world}.
\ryan{I took this from Sagot, so we should read
  the original.}

For a given part of speech, this quantity varies dramatically
over the languages of the world.  While the regular English verb
paradigm has three slots in our annotation, the Archi \ryan{\footnote{Archi is a Northeast Caucasian language, spoken in Dagestan, Russia.}} verb will have
thousands \cite{archi}.  However, does this make the Archi system more
complex?  In other words, is it more difficult to describe or to
learn?  Despite the plethora of forms, it is often the case that one
can regularly predict one form from another, indicating that few forms
actually have to be memorized for each lexeme.

\subsubsection{Integrative Complexity}
The second notion of complexity is {\bf integrative complexity} ({\bf
  i-complexity}), which measures how regular an inflectional system is
on the surface. Students of a foreign language will most
certainly have encountered the concept of an irregular verb.  Pinning
down a formal and workable cross-linguistic definition is non-trivial, but the
intuition that some inflected forms are regular and others irregular
dates back at least to \newcite[pp.~273--274]{bloomfield1933language},
\jason{did panini have this idea too?  how about, say, Latin grammarians?}
who famously argued that what makes a surface form regular is that it
is the output of a deterministic function. For an in-depth dissection of
the subject, see \newcite{stolz2012irregularity}.

\newcite{ackerman2013} build their definition of i-complexity on the information-theoretic notion of entropy \cite{Shannon1948}.  Their intuition is that a morphological system should be considered irregular to the extent that its forms are unpredictable.
They say, for example, that the nominative singular form is
unpredictable in a language if many verbs express it with suffix
\text{-{\it o}} while many others use \text{-$\emptyset$}.  In this
paper, we will propose an improvement to their entropy-based measure.

\Jason{we should say later that we measure e-complexity using tokens (is this actually good?), but i-complexity is measured over types.}

\Jason{we should cut the Kolmogorov discussion, as we discussed}

\subsection{The Low-Entropy Conjecture}\label{sec:tradeoffs}
The low-entropy conjecture, as formulated by \newcite[p.~436]{ackerman2013}, ``is the hypothesis that enumerative
morphological complexity is effectively unrestricted, as long as the
average conditional entropy, a measure of integrative complexity, is
low.'' In other words, morphological systems face a trade-off between
e-complexity and i-complexity: a system may be complex under either
metric, but not under both.  Indeed, Ackerman \& Malouf go so far as to
say that there need be no upper bound on e-complexity as long as the
i-complexity remains sufficiently low.

This line of thinking harks back to the equal complexity conjecture of Hockett, who stated: ``objective measurement is difficult, but impressionistically it would seem that the total grammatical complexity of any language, counting both the morphology and syntax, is about the same as any other'' \cite[pp.~180-181]{hockett1958course}.  Similar trade-offs have been found in other branches of linguistics (see \newcite{oh2015linguistic} for a review).  For example, there is a trade-off between rate of speech and syllable complexity \cite{pellegrino2011across}: this means that even though Spanish speakers utter many more syllables per second than Chinese, the overall information rate is quite similar as Chinese syllables carry more information (they mark tonality).

Hockett's {\em equal} complexity conjecture is controversial: languages such as Riau Indonesian seem low in complexity across morphology and syntax \cite{gil1994}.  This is why Ackerman and Malouf instead posit that a linguistic system has {\em bounded} complexity.  Their low-entropy conjecture says that the ``total'' complexity of a morphological system (e-complexity and i-complexity) must not be too high---though it can be low, as indeed it is in isolating languages like Chinese and Japanese.

\section{Entropic Integrative Complexity}\label{sec:paradigm}
In this section, we advocate for a probabilistic treatment
of paradigmatic morphology.
We assume that a language's inflectional morphology system
is a distribution over {\em possible} paradigms \cite{dreyer-eisner-2009,cotterell-peng-eisner-2015}.
For instance, knowing the 4-slot English verbal paradigm
means knowing a joint distribution over 4-tuples of surface forms,
\begin{equation}
  \pI(m_{\text{\sc lemma}}, m_{\text{\sc 3ps}}, m_{\text{\sc past}}, m_{\text{\sc gerund}})
\end{equation}
This is the ``base distribution'' from which each new word type's paradigm is assumed to have been sampled.  Each observed paradigm such as $\pI(\textit{run}, \textit{runs}, \textit{running}, \textit{ran})$ provides evidence of this distribution.  The fact that some paradigms are {\em used} more frequently than others (more tokens) does not mean that they have higher base probability under the morphological system $\pI$.  Rather, their higher usage is a semantic effect or simply a rich-get-richer effect \cite{dreyer-eisner-2011}.
\mans{We should clarify the subscript $I$ in $\pI$. Are we using it consistently here?}\jason{also, it probably shouldn't be italicized if it's not a variable}

We expect the base distribution to place low probability on implausible paradigms, e.g., $p(\textit{run}, \st{\textit{snur}}, \textit{running}, \st{\textit{nar}})$ is low---perhaps close to zero. Moreover, we expect the conditionals of this distribution to assign high probability to the result of applying regular processes, e.g., $\pI(\textit{sprint}, \textit{sprints}, \textit{sprinting} \mid \textit{sprinted})$ in English should be close to 1.  So should
  $\pI(\textit{wug}, \textit{wugs}, \textit{wugging} \mid
  \textit{wugged})$, where \textit{wug} is a novel word.  We note that
  $\pI$ (when smoothed) will have support over ${\Sigma^* \times
    \cdots \times \Sigma^*}$: it assigns positive probability to
  any $n$-tuple of strings. The model is thus capable of evaluating
  arbitrary wug-formations \cite{berko1958child}, including irregular ones.

So how do we relate $\pI$ to the i-complexity of a language?
Here, we again follow the spirit \ryan{\footnote{\newcite{ackerman2013} actually argued
  for the use of (averaged) conditional entropy. We will address
  this distinction head on in \cref{sec:comparison}.}} of
\newcite{ackerman2013} and argue that the entropy $H(\pI)$
is an appropriate measure, which is defined in our setting as
the joint entropy
\begin{equation}\label{eq:paradigm-entropy}
\!-\!\!\sum_{i=1\;}^n \! \sum_{\;\vec{m} \in (\Sigma^*)^n} \!\!\! \pI(m_1,\ldots,m_n)\log_2 \pI(m_1,\ldots,m_n).
\end{equation}

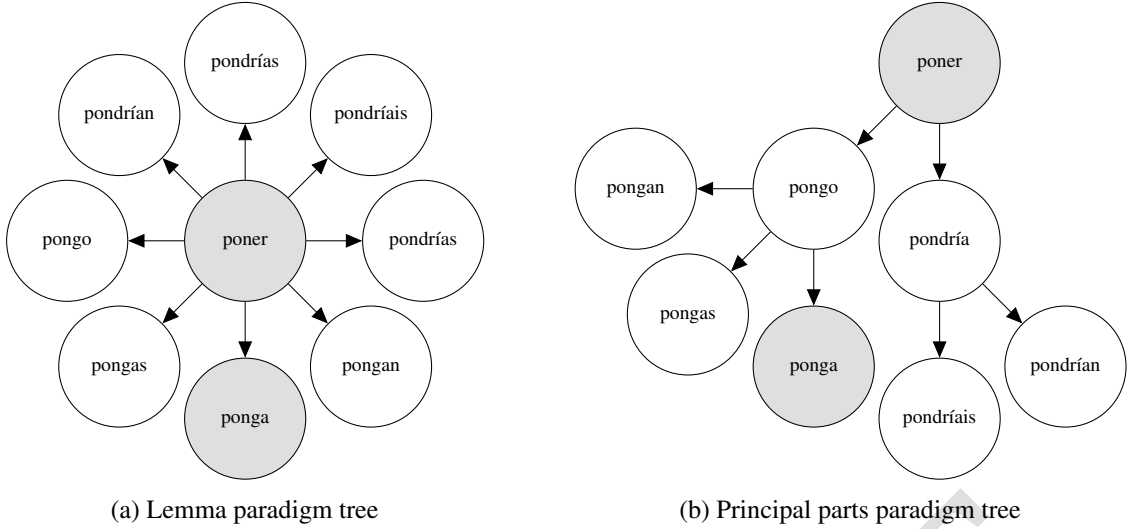
\begin{figure*}
  \begin{adjustbox}{width=2.0\columnwidth}
\begin{subfigure}[t]{\columnwidth}
  \centering
        \begin{tikzpicture}[scale=0.75, every node/.style={transform shape}]
          \node[obs] (lemma) {$\text{\myboxgrey{poner}}$} ; %
          \node[latent, left=of lemma, xshift=-0.0cm] (m1) {$\text{\mybox{pongo}}$} ; %
          \node[latent, below left=of lemma, yshift=-0.0cm] (m2) {$\text{\mybox{pongas}}$} ; %
          \node[obs, below=of lemma, yshift=0.0cm] (m3) {$\text{\myboxgrey{ponga}}$} ; %
          \node[latent, below right=of lemma, yshift=-0.0cm] (m4) {$\text{\mybox{pongan}}$} ; %
          \node[latent, right=of lemma, yshift=-0.0cm] (m5) {$\text{\mybox{pondr{\'i}as}}$} ; %
          \node[latent, above right=of lemma, yshift=-0.0cm] (m6) {$\text{\mybox{pondr{\'i}ais}}$} ; %
          \node[latent, above left=of lemma, yshift=-0.0cm] (m7) {$\text{\mybox{pondr{\'ian}}}$} ; %
          \node[latent, above=of lemma, yshift=-0.0cm] (m8) {$\text{\mybox{pondr{\'i}as}}$} ; %
          \edge {lemma} {m1, m2, m3, m4, m5, m6, m7, m8};
        \end{tikzpicture}
        \caption{Lemma paradigm tree}
        \label{fig:lemma}
\end{subfigure}
\begin{subfigure}[t]{\columnwidth}
  \centering
  \begin{tikzpicture}[scale=0.75, every node/.style={transform shape}]
    \node[obs] (m1) {$\text{\myboxgrey{poner}}$} ; %
    \node[latent, below left=of m1] (m2) {$\text{\mybox{pongo}}$} ; %
    \node[latent, below=of m1] (m3) {$\text{\mybox{pondr{\'i}a}}$} ; %
    \node[latent, below left=of m2] (m4) {$\text{\mybox{pongas}}$} ; %
    \node[obs, below=of m2] (m5) {$\text{\myboxgrey{ponga}}$} ; %
    \node[latent, left=of m2] (m6) {$\text{\mybox{pongan}}$} ; %
    \node[latent, below=of m3] (m7) {$\text{\mybox{pondr{\'i}ais}}$} ; %
    \node[latent, below right=of m3] (m8) {$\text{\mybox{pondr{\'i}an}}$} ; %
    \edge {m1} {m2, m3} ;
    \edge {m2} {m4, m5} ;
    \edge {m2} {m6} ;
    \edge {m3} {m7, m8} ;
  \end{tikzpicture}
  \caption{Principal parts paradigm tree}
  \label{fig:pp}
\end{subfigure}
\end{adjustbox}
\caption{Two potential directed graphical models for the paradigm completion task. The topology in
  (a) encodes the the network where all forms are predicted from the lemma. The topology in (b), on the other hand, makes it easier to predict forms given the others: \word{pongas} is predicted from \word{ponga},
  with which it shares a stem. Qualitatively, the structure learning algorithm discussed in \cref{sec:structure-learning} finds trees structured similarly to (b). }
\label{fig:topo}
\end{figure*}

\subsection{A Variational Upper Bound on Entropy}\label{sec:bound}
Lamentably, the paradigm entropy defined in \cref{eq:paradigm-entropy}
requires approximation. First, we do not actually know the true
distribution $\pI$.  Furthermore, even if we knew $\pI$, a sufficiently expressive
distribution would render direct computation intractable: it involves
$n$ nested sums over the infinite set $\Sigma^*$.  Thus, following \newcite{brown1992estimate}, we use a probability model to estimate an upper bound for the paradigm entropy.  Our starting point is a
well-known bound on the entropy of $\pI$,
\begin{equation}\label{eq:bound}
  H(\pI) \leq H(\pI, q),
  \end{equation}
where $q$ is any other distribution
over the same space
as $\pI$. In our setting, the cross-entropy $H(\pI, q)$
is defined as
\begin{equation}\label{eq:cross-entropy}
-\sum_{\;\vec{m} \in (\Sigma^*)^n} \! \pI(m_1,\ldots,m_n)\log  q(m_1,\ldots,m_n)
\end{equation}
The quality of the bound in \eqref{eq:bound}
depends on how close $q$ is to $\pI$, as measured by the KL-divergence $D(\pI~\mid\mid~q)$,
with equality in \eqref{eq:bound} if and only if $\pI = q$.

\paragraph{Choice of $q$.}
The bound in \cref{eq:bound} holds for any choice of $q$.  We cannot
practically search over all distributions to find the tightest
bound. Nevertheless, we can still find a reasonably good $q$ through
direct estimation of a probability model. Given a set of true morphological
paradigms ${\cal D}_{\text{train}}$ drawn from $\pI$, we can fit our probability model
$q$ in any reasonable way, for example by (locally) maximizing the
log-likelihood \jason{why not smooth?}
\begin{equation}\label{eq:mle}
  \sum_{\vec{m} \in {\cal D}_{\text{train}}} \log q(m_1, \ldots, m_n),
\end{equation}
This is equivalent to seeking the tightest bound \eqref{eq:bound} achievable by any
distribution $q$ in a parametric family ${\cal Q}$. We discuss our specific
choice of ${\cal Q}$ in \cref{sec:generative} below.

\paragraph{Estimate of i-complexity.}  Having chosen $q$, we
can estimate $H(\pI, q)$ using a separate held-out
sample:\footnote{Both ${\cal D}_{\text{test}}$ and ${\cal
    D}_{\text{train}}$ should be sampled from $\pI$.  They
  should be independent or disjoint samples for \cref{eq:approx} to be a good estimate of the cross-entropy.}
\begin{align}\label{eq:approx}
  H(\pI, q) &\approx -\frac{1}{d}
  \sum_{\vec{m} \in {\cal D}_{\text{test}}} \log q(m_1, \ldots, m_n)
\end{align}
where $d = |{\cal D}_{\text{test}}|$.  This can be computed as long as
we can evaluate $q$, and the estimate converges to $H(\pI,q)$ as the
test sample size $d \rightarrow \infty$.  We return this estimate as
our practical approximation of the desired e-complexity $H(\pI)$.

\section{A Generative Model of the Paradigm}\label{sec:generative}

To fit $q$, we need a tractable parametric family ${\cal Q}$ of joint
distributions over paradigms.  To define ${\cal Q}$, we follow
\newcite{cotterell-sylakglassman-kirov:2017:EACLshort} and arrange the
$n$ slots \jason{but what if some slots are empty?  We promised in
  \cref{fn:not-plat} that they could be.}
into a tree-structured Bayesian network (a directed graphical
model).  For a given tree ${\cal T}$, we have the function
$\text{pa}_{\cal T}(i)$, which returns the parent of the $i^\text{th}$
cell or the empty string if the $i^\text{th}$ cell is the
root. \jason{okay that I said ``the empty string'' instead of
  ``nothing''?  I did it again below.} We show two possible tree
structures for Spanish verbs in \cref{fig:topo}. Now, we may write a
particular element of ${\cal Q}$---a factored joint distribution over
all $n$ forms---as
\begin{equation}\label{eq:bayesnet}
  \qtheta(m_1, \ldots, m_n) = \prod_{i=1}^n \qtheta(m_i \mid m_{\text{pa}_{\cal T}(i)}).
\end{equation}
where $\vtheta$ represents the parameter vector of the Bayesian
network.  We specifically model all of the conditional probabilities
in \eqref{eq:bayesnet} using a neural sequence-to-sequence
model with parameters $\vtheta$, as described in \cref{sec:kann} below.
As our distribution $\qtheta$ is a smooth function of
$\vtheta$, we can maximize \eqref{eq:mle} via gradient-based
optimization, as outlined in \cref{sec:experimental-details}.

\subsection{Neural Sequence-to-Sequence Model}\label{sec:kann}

The state of the art in morphological reinflection \cite{kann-schutze:2016} uses an LSTM-based sequence-to-sequence model \cite{DBLP:conf/nips/SutskeverVL14} with attention \cite{DBLP:journals/corr/BahdanauCB14}.
The idea is to model reinflection as ``translation'' of an input character
sequence, with a description of the desired output slot appended to the
input sequence in the form of special characters.  For example,
to

For example, in German, consider the mapping
from the nominative singular form \word{Hand}
to the nominative plural form \word{H{\"a}nde}.
This is encoded with the source string {\footnotesize {\tt H a n d IN=NOM IN=SG OUT=NOM OUT=PL} }
and target string {\footnotesize {\tt H {\" a} n d e}}.
If the slot realizes multiple feature bundles, we append
each of them to the input source string.
This encoding may be suboptimal, as it throws away which
features belong to which bundles.
\jason{what do
  you do if a slot realizes multiple feature bundles, as discussed in \cref{sec:word-based-morphology}?}
This is similar to the encoding in
\newcite{kann-schutze:2016}, and allows the same LSTM with parameters
$\vtheta$ to be reused at each factor of \eqref{eq:bayesnet}.  Different
factors $\qtheta(m_i \mid m_j)$
are distinguished only by the fact that the
morphological tags for slots $i$ and $j$ are appended to the
input string before the LSTM is applied to it.

\subsection{Structure Learning}\label{sec:structure-learning}
Which  tree over the $n$ slots is optimal? It is not clear {\em a-priori} how to
arrange the slots in a paradigm such that their predictability is
maximized. For instance, consider the irregular Spanish verb
\word{poner}, we may want to predict its present subjunctive forms,
e.g., \word{ponga}, \word{pongas} and \word{ponga}, from another form
that shares the same stem, e.g., \word{ponga}---this maximizes
predictability in that we no longer have to account for the irregular
present subjective stem change. Our goal, however, is to select
the optimal tree for the data, rather than pre-specified linguistic
knowledge of the language.

In graph-theoretic terms, we choose the highest-weighted directed
spanning tree over $n$ vertices, as found by the algorithm of
\newcite{edmonds1967optimum}.  The weight of a candidate tree is the
sum of all its edge weights and the weight of its root vertex, where
we define the weight of a candidate edge to $m_i$ from $m_j$ as
$\frac{1}{d} \sum_{\vec{m} \in {\cal D}_{\text{dev}}} \log q(m_i \mid
m_j)$,
and define the weight of vertex $m_i$ as
$\frac{1}{d} \sum_{\vec{m} \in {\cal D}_{\text{dev}}} \log q(m_i \mid
\text{empty string})$, where ${\cal D}_{\text{dev}}$ is a set of
development paradigms.  In each case, $q$ is a sequence-to-sequence
model trained on
${\cal D}_{\text{train}}$, so computing these $n^2$ weights requires
us to train $n^2$ sequence-to-sequence models.  Under this scheme, the
weight of a candidate tree is the log-likelihood
$\frac{1}{d} \sum_{\vec{m} \in {\cal D}_{\text{dev}}} \log
q(m_1,\ldots,m_n)$
of a model whose structure is given by the tree and whose conditional
distributions are given by these trained $q$ distributions.
Recall that our estimate of $H(p,q)$ is the same, but evaluated on
${\cal D}_{\text{test}}$ (\cref{eq:approx}).

In fact, as in \cref{sec:kann}, we train only a single shared
LSTM-based sequence-to-sequence model to perform all $n^2$
transductions.  Once we have selected the tree, we could retrain the
model to focus on only the $n$ transductions actually required by the
tree, but our present experiments do not retrain.

\begin{table}
  \begin{adjustbox}{width=\columnwidth}
    \begin{tabular}{l llll llll} \toprule
      &  \multicolumn{4}{c}{{\sc singular}} & \multicolumn{4}{c}{{\sc plural}} \\ \cmidrule(l){1-1} \cmidrule(l){2-5} \cmidrule(l){6-9}
    {\sc class} & {\sc nom} & {\sc gen} & {\sc acc} & {\sc voc} & {\sc nom} & {\sc gen} & {\sc acc} & {\sc voc} \\ \midrule
    1 & -\word{os}   & -\word{u}    & -\word{on}   & -\word{e}     & -\word{i}    & -\word{on}    & -\word{us}   & -\word{i} \\
    2 & -\word{s}    & -$\emptyset$ & -$\emptyset$ & -$\emptyset$  & -\word{es}   & -\word{on}    & -\word{es}   & -\word{es} \\
    3 & -$\emptyset$ & -\word{s}    & -$\emptyset$ & -$\emptyset$  & -\word{es}   & -\word{on}    & -\word{es}   & -\word{es} \\
    4 & -$\emptyset$ & -\word{s}    & -$\emptyset$ & -$\emptyset$  & -\word{is}   & -\word{on}    & -\word{is}   & -\word{is} \\
    5 & -\word{o}    & -\word{u}    & -\word{o}    & -\word{o}     & -\word{a}    & -\word{on}    & -\word{a}    & -\word{a} \\
    6 & -$\emptyset$ & -\word{u}    & -$\emptyset$ & -$\emptyset$  & -\word{a}    & -\word{on}    & -\word{a}    & -\word{a} \\
    7 & -\word{os}   & -\word{us}   & -\word{os}   & -\word{os}    & -\word{i}    & -\word{on}    & -\word{i}    & -\word{i} \\
    8 & -$\emptyset$ & -\word{os}   & -$\emptyset$ & -$\emptyset$  & -\word{a}    & -\word{on}    & -\word{a}    & -\word{a} \\ \bottomrule
  \end{tabular}
  \end{adjustbox}
  \caption{Structuralist analysis of Modern Greek nominal inflection classes.
 \cite{ralli1994feature,ralli2002role}.}
  \label{tab:greek}
\end{table}

\section{A Methodological Comparison to \newcite{ackerman2013}}\label{sec:comparison}

Our formulation of the low-entropy principle differs somewhat
from \newcite{ackerman2013}. We highlight the differences.

\paragraph{Heuristic Approximation to $\pI$.}
\newcite{ackerman2013} first construct what we regard as a heuristic
approximation to the joint distribution $\pI$ over forms in a
paradigm. They first provide a structuralist decomposition of words
into their constituent morphemes. Then, they consider a distribution
$r(m_i \mid m_j)$ that builds new forms by swapping morphemes. In
contrast to our neural sequence-to-sequence approach, this
distribution unfortunately does {\em not} have support over $\Sigma^*$
and, thus, cannot consider changes other than substitution of affixes.
\Jason{Restore text about irregularity? Ryan originally wrote earlier:
  ``For example, it is harder to predict forms that are remnants of
  no-longer-productive ablaut processes (English
  $\word{sing} \mapsto \word{sang}$,
  $\word{swim} \mapsto \word{swam}$) or that are suppletive (English
  $\word{go} \mapsto \word{went}$).''  \newcite{ackerman2013} can't
  handle these examples, and we should also acknowledge that these
  small pockets of irregularity will actually get drowned out in a
  long list.  Our measure, like theirs, is more interested in {\em
    systematic} unpredictability, e.g., uncertainty about the
  conjugation class or which of several idiosyncratic endings is
  used.}

As concrete example of $r$, consider \cref{tab:greek}'s Modern Greek
example from \newcite{ackerman2013}.  The conditional distribution
$r(m_{\text{{\sc gen};{\sc sg}}} \mid m_{\text{{\sc acc};{\sc pl}}} =
\text{-{\it i}})$,
over genitive singular forms is peaked since there is exactly one possible
transformation: substituting {\em -us} for {\em -i}. This is not always
the case for Modern Greek, \newcite{ackerman2013} estimated that
$r(m_{\text{{\sc nom};{\sc sg}}} \mid m_{\text{{\sc acc};{\sc pl}}} =
\text{-{\it a}})$
swaps {\em -a} for $\emptyset$ with probability $\sfrac{2}{3}$ and for
{\em -o} with probability $\sfrac{1}{3}$. We reiterate that no other
transformation would be possible, \jason{do you mean they are
  deterministic?} e.g., swapping {\em -a} for {\em -es} or mapping it
to some arbitrary form such as {\em foo}.

\paragraph{Average Conditional Entropy.}
The second difference is their reliance on the pair-wise conditional entropy between
two cells. That is, they argue for the quantity
\begin{equation}
H(i \mid j) = -\!\!\!\! \sum_{m_i \in \Sigma^*} r(m_i) \log r(m_i \mid m_j),
\end{equation}
where $m_j$ is a given form. (We have written the sum over $\Sigma^*$, but as $r$ has finite support,
in practice one only has to consider the possible reinflections of $m_j$ the annotation of the data admits.) The entropy of an entire paradigm, is then the average
conditional entropy:
\begin{equation}\label{eq:average-conditional-entropy}
  \frac{1}{n^2 - n} \sum_{i=1}^n \sum_{j=i+1}^n H(i \mid j).
\end{equation}

\subsection{Critique of \newcite{ackerman2013}}
Now, we offer a critique of \newcite{ackerman2013} on three  points: (i)
different linguistic theories may offer different results,
(ii) there is no principled manner to handle morphological
irregularity, and (iii) average conditional entropy overestimates
the i-complexity in comparison to joint entropy. We discuss each in turn.

\paragraph{Theory-dependent Entropy.}
We consider a classical example from English morpho-phonology that
demonstrates the dependence of paradigm entropy on the specific
analysis chosen. In regular English plural formation, the speaker has
three choices: [z], [s] and [\textipa{1}z]. Here are two potential analyses. One the
one hand, we may treat this as a case of pure allomorphy with three
potential, unrelated suffixes. Under such an analysis, the entropy
will reflect the empirical distribution: roughly, $\sfrac{1}{4} \log
\sfrac{1}{4} + \sfrac{3}{8} \log \sfrac{3}{8} + \sfrac{3}{8} \log
\sfrac{3}{8} \approx 1.56127$\ryan{(DISCUSS WHERE I GOT THESE NUMBERS)}. On the other
hand, if we assume a unique underlying affix /z/, which is attached
and then converted to either [z], [s] or [\textipa{1}z] by an application of perfectly regular phonology, this part of the morphological system of English has entropy of
0---one choice. See \newcite[p.72]{kenstowicz1994} for a discussion of these alternatives
from a theoretical standpoint. Note that our goal is
not to advocate for one of these analyses, but merely to suggest that
\newcite{ackerman2013}'s quantity is analysis-dependent. In contrast,
our approach is theory-agnostic in that we jointly learn string-to-string
transformations, reminiscent of a-morphorous morphology \cite{anderson1992morphous}, and thus our
(approximation to) paradigm entropy does not suffer this drawback.
Indeed, our assumptions
are limited---recurrent neural networks are universal algorithm
approximators.  It has been shown that there exists a finite RNN that can
compute any computable function
\cite{siegelmann1991turing,siegelmann1995computational}. Thus, the
only true assumption we make of morphology is mild: we assume it is Turing-computable;
that language is Turing-computable is a fundamental tenet of cognitive science
\cite{mcculloch1943logical,sobel2013cognitive}.

\paragraph{Morphological Irregularity.}
A second problem with \newcite{ackerman2013} is its
treatment of irregularity, e.g., cases of suppletion.
As far as we can tell, the model is incapable of evaluating
cases of morphological suppletion unless they are explicitly
encoded in the model. Consider, again, the case
of the English suppletive past tense form \word{went}---
if your analysis of the English base is effectively a distribution
of the choices add [d], add [t] and [\textipa{1} d], you will
assign probability 0 to {\em went} as the past tense of {\em go}.
We highlight the importance of this point because suppletive forms are certainly
very common in academic English: the plural of \word{binyan} is \word{binyanim}
and the plural of \word{lemma} is \word{lemmata}. It is unlikely that
native English speakers have even a partial model of Hebrew and Greek, respectively, nominal morphology in their heads---a more plausible scenario is simply
that these forms are learned by rote. As speakers and hearers are capable
of producing and analyzing these forms, we should demand the same capacity
of our models. We note that these restrictive assumptions are
relatively common in the literature, e.g., \newcite{allen2015learning}'s sublexical learner is likewise incapable of placing probability mass on irregulars.\footnote{We point out that in the computer science literature, it is far
  more common to construct distributions with support over $\Sigma^*$ \cite{paz2003probabilistic,bouchard-EtAl:2007:EMNLP-CoNLL2007,dreyer-smith-eisner:2008:EMNLP,cotterell-peng-eisner:2014:P14-2}, all of which are perfectly capable of evaluating arbitrary formations.}
\Mans{We might be tight on space, but another closely related complaint is that A\&M not only are unable to handle irregularity, but also draw an arbitrary line between `regular' morphology (what they calculate with) and `irregular' morphology (which they skip) which we, arguably, aren't guilty of.}
\paragraph{Average Conditional Entropy versus Joint Entropy.}
Finally, we take issue with the formulation of paradigm entropy as
average conditional entropy, as exhibited in \cref{eq:average-conditional-entropy}. For one, it
does not correspond to the entropy of any one joint distribution as
the product of the conditionals does not yield the joint; this denies
the quantity a clean mathematical interpretation. Second, it is
Priscian \cite{robins2013} in its analysis
in that any form can be generated from any other, which, in practice, will
cause it to overestimate the i-complexity of a morphological system. Consider
the German dative plural
\word{H{\"a}nden} (from the German \word{Hand} ``hand''). Predicting
this form from the nominative singular \word{Hand} is difficult, but
predicting it from the nominative plural \word{H{\"ande}} is trivial:
just add the suffix \word{-n}. In \newcite{ackerman2013}'s
formulation, $r(\textit{H{\"a}nden} \mid \textit{Hand})$ and
$r(\textit{H{\"a}nden} \mid \textit{H{\"a}nde})$ both contribute to
the paradigm's entropy with the former raising the quantity. We
believe this is suboptimal and, as we have shown in
\cref{sec:generative}, an entropy-based formulation of morphological
complexity need not have this property, i.e., only one of the two
conditional entropies must count towards the final entropy of the
paradigm, as is the case in the minimum spanning aborescence.

\section{Experiments}
The crux of our experimentation is simple: we will
plot e-complexity versus i-complexity over as many languages as possible, and
then devise a numerical test of whether the low-entropy conjecture appears to hold.

\subsection{Data and UniMorph Annotation}\label{sec:unimorph}

At the moment, the largest source of annotated full paradigms is the
UniMorph dataset \cite{sylakglassman-EtAl:2015:ACL-IJCNLP}, which contains
data that have been extracted from Wiktionary, as well as other morphological lexica and analyzers,
and then converted into a universal format. A partial subset
of Unimorph has been used in the running of the SIGMORPHON-CoNLL
2017 shared task on morphological inflection generation \cite{cotterell-conll-sigmorphon2017}.

We use verbal paradigms from 23 typologically diverse languages, and
nominal paradigms from 31 typologically diverse languages.  These are
the UniMorph languages that contain at least 500 distinct verbal or
nominal paradigms.\footnote{UniMorph currently contains 51 languages
  altogether.  Over 100 additional languages are in preparation, which
  will allow even larger-scale experiments in the future.}  As the
neural methods require a large set of annotated training examples to
achieve high performance, it is difficult to use them in a
lower-resource scenario.

\paragraph{Empirically Measuring i-Complexity.}  Here we follow the
procedure from \cref{sec:bound} and \cref{sec:generative}. That is, we
partition the available paradigms into training, development and test
sets.  We train the factors of our generative model (\cref{sec:kann}) on the
training set, selecting among potential model structures on the
development set using Edmonds's algorithm
(\cref{sec:structure-learning}), and then evaluate i-complexity on the
unseen test set (\cref{sec:bound}).  Using held-out data in this way
gives a fair estimate of the actual predictability (i-complexity) of
the paradigms, which is why it is standard practice on most common NLP
tasks,
though less common in quantitative approaches to linguistic theory.

\begin{figure*}[h!]
  \begin{subfigure}[t]{\columnwidth}
    \centering
      \includegraphics[width=\columnwidth]{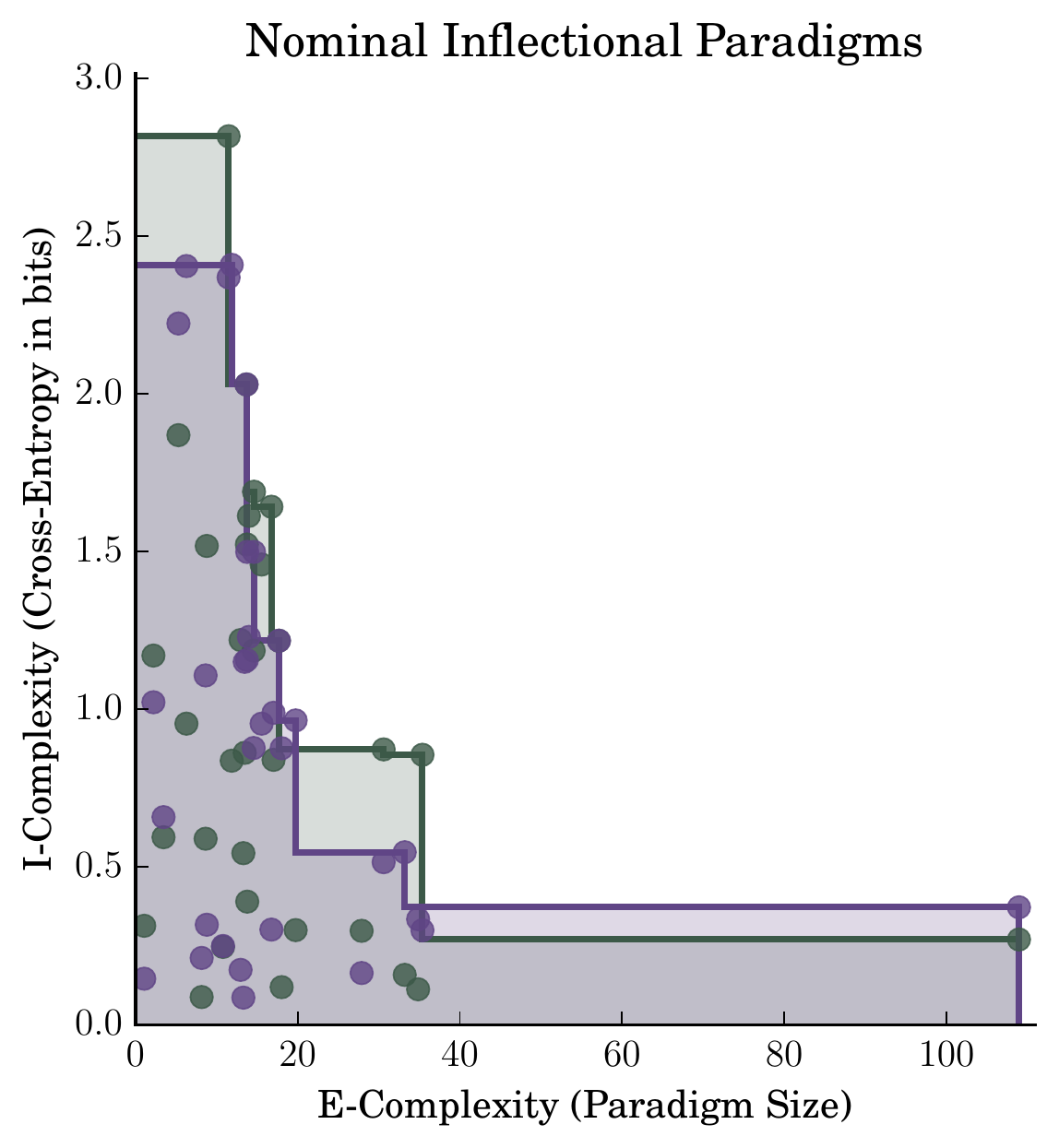}
    \end{subfigure}
  \begin{subfigure}[t]{\columnwidth}
    \centering
      \includegraphics[width=\columnwidth]{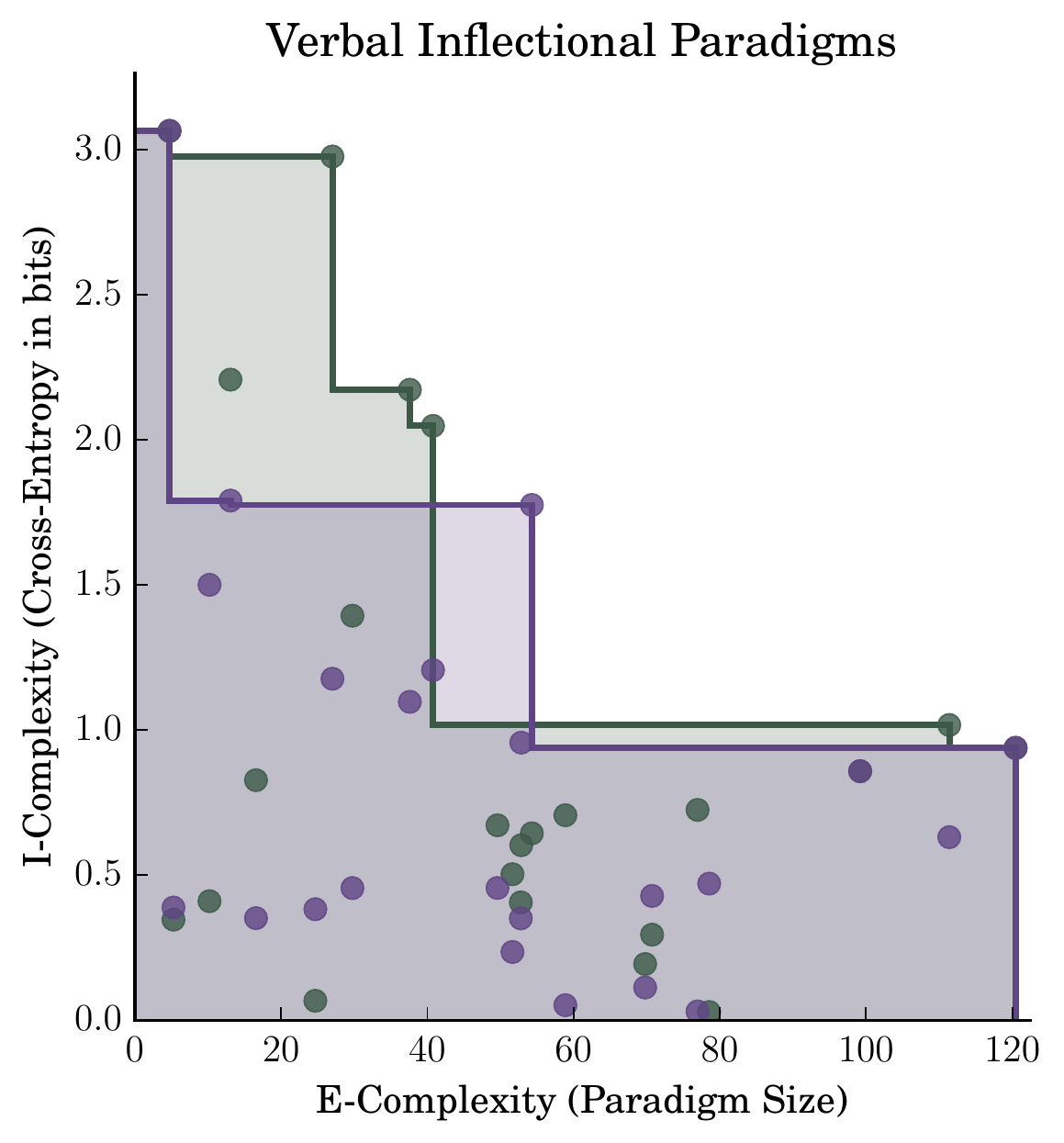}
    \end{subfigure}
  \caption{The $y$-axis is cross-entropy, an approximation to
  the paradigm and entropy and a measure of i-complexity. The $x$-axis
  is the size of the paradigm, a measure of e-complexity. Both of these graphs
  overlay purple and green points, as discussed in \cref{sec:experimental-details}.
  For concreteness, the purple points are those models trained with the same number of paradigms
  observed at training time across languages and the green points are those models trained with the same
  number of slot-to-slot mappings observed at training time.  The
  purple curve is the Pareto curve for the purple points, and the area
  under it is shaded in purple; similarly for green.
  }
  \label{fig:graph}
\end{figure*}

\paragraph{Empirically Measuring E-Complexity.}
The measurement of the e-complexity in this scheme is relatively
straightforward. Following Ackerman and Malouf, we simply count
the number of slots in the paradigm (for nouns or for verbs as
appropriate).\footnote{Occasionally forms are missing for a given
  lexeme, so we take the maximum size of all the nominal paradigms or
  all the verbal paradigms.}

\begin{table}[ht!]
  \centering
  \begin{adjustbox}{width=\columnwidth}
    \begin{tabular}{lllll} \toprule
     & \multicolumn{2}{c}{Nouns} & \multicolumn{2}{c}{Verbs} \\ \cmidrule(lr){1-1} \cmidrule(lr){2-3} \cmidrule(lr){4-5}
    Language & $|\pi|$ & $H(\pI, \qtheta)$ & $|\pi|$ & $H(\pI, \qtheta)$  \\ \midrule
    Arabic           & 112 & 0.44 & 36 & 0.21 \\
    Armenian         & -- & -- & 34 & 0.23 \\
    Bulgarian        & 52 & 0.666 & 9 & 0.22 \\
    Catalan          & 53 & 0.24 & -- & -- \\
    Czech	     & -- & -- & 14 & 0.61 \\
    Danish           & -- & -- & 6 & 1.67 \\
    Dutch            & 16 & 0.24 & -- & --  \\
    English	     & 5 & 0.27 & 2 & 0.10  \\
    Estonian         & -- & -- & 30 &  0.38 \\
    Faroese	     & 14 & 1.24 & 16  & 0.21 \\
    Finnish	     & -- & -- &  28 & 0.11 \\
    French           & 49 & 0.32 & -- & -- \\
    Georgian         & -- & -- & 19 & 0.61 \\
    German	     & 29 & 0.32 & 8 & 0.77 \\
    Hungarian        & 59 & 0.04 & 34 & 0.38 \\
    Icelandic        & -- & -- & 16 & 0.66 \\
    Irish	     &  -- & -- & 13 & 0.06 \\
    Latin	     & 100    & 0.59 & 12 & 0.12 \\
    Latvian	        & --  & -- & 12 & 0.12 \\
    Lithuanian	        & --  & -- & 14 & 1.04 \\
    Lower Sorbian       & --  & -- & 18 & 0.84 \\
    Macedonian          & 79  & 0.33 & 11 & 0.17 \\
    Northern Kurdish    & --  & --   & 20  & 0.67 \\
    Northern Sami       & 54  & 1.23 & 13  & 0.80 \\
    Norwegian Bokm\aa l & 5   & 2.12 & 3   & 0.71  \\
    Norwegian Nynorsk   & --  & --   & 3   & 0.46 \\
    Polish	        & --  & --   & 14  & 0.80 \\
    Romanian            & 37  & 0.76 & 6   & 1.54 \\
    Russian             & 25  & 0.27 & 12  & 1.67 \\
    Serbo-Croatian      & 70  & 0.08 & 14  & 1.41 \\
    Slovak	        & --  & --   & 12  & 1.64 \\
    Slovenian	        & --  & --   & 18  & 0.69 \\
    Spanish             & --  & --   & 70  & 0.30 \\
    Swedish	        & 11  & 1.04 & 8 & 0.15 \\
    Turkish	        & 120 & 0.65 & 108 & 0.26 \\
    Ukrainian           & --  & --   & 14 & 0.85 \\
    \end{tabular}
  \end{adjustbox}
  \caption{The set of 36 languages used in our experiment with their paradigm size $|\pi|$ and our i-complexity measure $H(\pI, \qtheta)$ (green training scheme; purple scheme, omitted for space.)}
  \label{tab:results}
\end{table}

\subsection{Experimental Details}\label{sec:experimental-details}
For the i-complexity experiments, we split the full set of UniMorph
nominal paradigms into train, development, and test sets as
follows. We held out at random 50 full paradigms for the development
set, and 50 others for the test set. To form the development and test sets,
we include all pairwise mappings between inflected forms in each of
the 50 paradigms, except the identity mapping.

We sampled ${\cal D}_{\text{train}}$ from the remaining data.
We tried two ways of doing this, which deal differently with the fact that different languages
have a different number of slots per paradigm.  Both regimes seemed
reasonable, so we tried it both ways to confirm that the choice did
not affect the qualitative results.

\paragraph{Equal Number of Paradigms (Purple).}
In the first regime, ${\cal D}_{\text{train}}$ (for each language)
contains 600 paradigms chosen from the non-held-out data.  We trained
the reinflection model in \cref{sec:structure-learning} on all $n^2$
mappings from these paradigms. Henceforth, we will abbreviate this
training regime to the purple scheme.

\paragraph{Equal Number of Pairs (Green).}
In the second regime, we trained the reinflection model in
\cref{sec:structure-learning} on 60,000 $(m_i,m_j)$ or
$(m_i,\text{empty string})$ pairs sampled without replacement from the
non-held-out paradigms.\footnote{For a few languages, fewer than
  60,000 pairs were available, in which case we used all pairs.}
This matches the amount of training data, but may disadvantage languages with large
paradigms, since the reinflection model will see fewer examples of any individual
mapping between paradigm slots.
Henceforth, we will abbreviate this
training regime to the green scheme.

\paragraph{Model and Training Details.}
We use the OpenNMT toolkit \cite{klein2017opennmt}. We largely follow
the recipe given in \newcite{kann-schutze:2016}, the winning
submission on the 2016 SIGMORPHON shared task for inflectional
morphology. Accordingly, we use a character embedding size of 300, and
100 hidden units in both the encoder and decoder. Our gradient-based
optimization method was AdaDelta \cite{zeiler2012adadelta} with a
minibatch size of 80. We trained for 20 epochs and select the test
model based on the performance on the development set. We decoded with
beam search with a beam size of 12.

\section{Results and Analysis}

Our results are listed in \cref{tab:results} and plotted in
\cref{fig:graph}, where each dot represents a language.  We saw little
difference between the green and the purple training schemes, though
it was not clear \textit{a-priori} that this would be the case.

The plots appear to show a clear trade-off between i-complexity and
the e-complexity.  We now provide quantitative support for this
impression, by constructing a statistical significance test.

Visually, Ackerman and Malouf's low-entropy conjecture boils
down to the claim that languages cannot exist in the upper right-hand
corner of the graph, i.e., they cannot have both high e-complexity and
high i-complexity. In other words, the upper-right hand corner of the
graph is ``emptier'' than it would be by chance.

How can we quantify this?  The {\bf Pareto curve} for a multiobjective
optimization problem shows, for each $x$, the maximum value $y$ of the
second objective that can be achieved while keeping the first
objective $\geq x$ (and vice-versa).  This is shown in
\cref{fig:graph} as a step curve, showing the maximum i-complexity
$y$ that was actually achieved for each level $x$ of e-complexity.
This curve is the tightest non-increasing function that upper-bounds all of
the observed points: we have no evidence from our sample of languages
that any language can appear above the curve.

We say that the upper right-hand corner is ``empty'' to the extent
that the area under the Pareto curve is small.  To ask whether it is
indeed emptier than would be expected by chance, we perform a nonparametric
permutation test that destroys the claimed correlation between the
e-complexity and i-complexity values.  From our observed points
$\{(x_1,y_1), \ldots, (x_m,y_m)\}$, we can stochastically construct a
new set of points
$\{(x_1,y_{\sigma(1)}), \ldots, (x_m,y_{\sigma(m)})\}$ where $\sigma$
is a permutation of $1,2,\ldots,m$ selected uniformly at random.  The
resulting scatterplot is what we would expect under the null
hypothesis of no correlation.  Our $p$-value is the probability that
the new scatterplot has an even emptier upper right-hand corner---that
is, the probability that the area under the null-hypothesis Pareto
curve is $\leq$ the area under the actually observed Pareto curve.  We
estimate this probability by constructing 10,000 random scatterplots.

In the purple training scheme, we find that the upper right-hand corner
is significantly empty, with $p < 0.017$ and $p < 0.045$ for the verbal
and nominal paradigms, respectively. In the green training scheme, we find
that the upper right-hand corner is significantly empty with $p < 0.042$
and $p < 0.034$ in the verbal and nominal paradigms, respectively.

\section{Future Directions}

\paragraph{Learnability.}
Ackerman \& Malouf's hypothesis is an interesting
starting point for future work.  It seems to be implicitly motivated
by the notion that naturally occurring languages must be learnable.
In other words, the intuition is that languages with large paradigms
need to be regular {\em overall}, because in such a language, the {\em
  average} word type is observed too rarely for a learner to memorize
an irregular surface form for it.  Yet even in such a language, {\em
  some} word types are frequent, because some lexemes and some slots
are especially useful.  Thus, if learnability of the lexicon is indeed
the driving force,\footnote{Rather than, say, description length of
  the lexicon \cite{rissanen94}.} then we should make the
finer-grained conjecture that irregularity (unpredictability) will be
better tolerated for the more frequently observed word types,
regardless of paradigm size.  Better yet, we should directly
investigate whether naturally occurring inflectional systems
are more learnable (at least by machine learning algorithms)
than would be expected by chance.  This is what one would predict
if languages are shaped by natural selection or, more plausibly, by
noisy transmission from each generation to the next
\cite{hare1995learning,smith2008introduction}.

\Jason{Our notion of i-complexity as a single entropy may be naive.  We've assumed [discuss earlier?] that the paradigms of a language are sampled IID from the same distribution, so that we can measure its entropy.  But in fact, high-frequency types are more likely to be irregular, meaning that there's not a single distribution---the distribution is conitioned on frequency.  Furthermore, this may be explained by a tendency for rarer types to regularize over time, so the entropy may differ depending on when you look, and the strength of the regularization tendency might also be a measure of the language's regularity.}

\Jason{In the future, we might want to study whether children's morphological systems increase in i-complexity as they approach the adult system.  Ryan wrote: ``Interestingly, this definition of i-complexity could also explain certain issues in first language acquisition, where children often overregularize \cite{pinker1988language}: they impose the regular pattern on irregular verbs, producing forms like \word{\st{runned}} instead of \word{ran}: children may initially posit [Jason: or at least arrive first at] an inflectional system with lower i-complexity, before converging on the true system, which has higher i-complexity.''}

\paragraph{Moving Beyond the Forms.}
The complexity of morphological inflections is only a small bit
of the larger question of morphological typology.
We have left many bits unexplored. In the realm of morphology,
for instance, we have conditioned on the morpho-syntactic
feature bundles. Ideally, we would like to explain
the underlying mechanisms that give rise to these
feature-bundles and the distinctions they make.

In addition, our current treatment depends upon a paradigmatic
treatment of morphology.  While viewing inflectional morphology as
paradigmatic is not controversial, derivational morphology is still
often viewed as syntagmatic. Can we discover a quantitative
formulation of derivational complexity?
We note that paradigmatic treatments of derivational morphology have been offered: see \newcite{cotterell-EtAl:2017:EMNLP2017} for a computational perspective
and the references therein for theoretical positions and arguments.

\section{Conclusions}

We have provided a clean mathematical formulation of enumerative and integrative
complexity of inflectional systems, using tools from generative
modeling and deep learning. With a empirical study on 36 typologically
diverse languages, we have shown that there is a Pareto-style
trade-off between e-complexity and i-complexity in morphological
systems.  In short, this means that morphological systems can either
mark a large number of morpho-syntactic distinctions, as Finnish,
Turkish and other agglutinative and polysynthetic languages do, or
they may have a high-level of unpredictability, i.e., irregularity.

This trade-off is a bit different than other trade-offs in linguistic typology, in that a language is under no obligation to be morphologically rich---it may have low e-complexity and i-complexity.  \newcite{carstairs2010evolution} has pointed out that languages need not have morphology at all, though they must have phonology and syntax.

Interestingly, NLP has largely focused on e-complexity.  Our community views a language as morphologically complex if it has a profusion of unique forms, even if they are very predictable. The reason is probably our habit of working at the word-level, so that all forms not found in the training set are out-of-vocabulary. However, as NLP moves to the character-level, we will need other definitions of morphological richness.  A language like Hungarian with almost perfectly predictable morphology may be easier to process than a language like German with an abundance of irregularity.

\bibliography{pareto}

\begin{thebibliography}{}

\bibitem[\protect\citename{Ackerman and Malouf}2013]{ackerman2013}
Farrell Ackerman and Robert Malouf.
\newblock 2013.
\newblock Morphological organization: {T}he low conditional entropy conjecture.
\newblock {\em Language}, 89(3):429--464.

\bibitem[\protect\citename{Allen and Becker}2015]{allen2015learning}
Blake Allen and Michael Becker.
\newblock 2015.
\newblock Learning alternations from surface forms with sublexical phonology.
\newblock {\em Unpublished manuscript, University of British Columbia and Stony
  Brook University. Available as lingbuzz/002503}.

\bibitem[\protect\citename{Anderson}1992]{anderson1992morphous}
Stephen~R. Anderson.
\newblock 1992.
\newblock {\em A-morphous Morphology}, volume~62.
\newblock Cambridge University Press.

\bibitem[\protect\citename{Aronoff}1976]{aronoff1976word}
Mark Aronoff.
\newblock 1976.
\newblock {\em Word Formation in Generative Grammar}.
\newblock Number~1 in Linguistic Inquiry Monographs. MIT Press, Cambridge, MA.

\bibitem[\protect\citename{Baerman \bgroup et al.\egroup
  }2015]{baerman2015understanding}
Matthew Baerman, Dunstan Brown, and Greville~G. Corbett.
\newblock 2015.
\newblock Understanding and measuring morphological complexity: {A}n
  introduction.

\bibitem[\protect\citename{Baerman}2015]{baerman2015oxford}
Matthew Baerman.
\newblock 2015.
\newblock {\em The Oxford Handbook of Inflection}.
\newblock Oxford Handbooks in Linguistic.
\newblock Part II: Paradigms and their Variants.

\bibitem[\protect\citename{Bahdanau \bgroup et al.\egroup
  }2015]{DBLP:journals/corr/BahdanauCB14}
Dzmitry Bahdanau, Kyunghyun Cho, and Yoshua Bengio.
\newblock 2015.
\newblock Neural machine translation by jointly learning to align and
  translate.
\newblock In {\em ICLR}.

\bibitem[\protect\citename{Berko}1958]{berko1958child}
Jean Berko.
\newblock 1958.
\newblock The child's learning of {E}nglish morphology.
\newblock {\em Word}, 14(2-3):150--177.

\bibitem[\protect\citename{Bloomfield}1933]{bloomfield1933language}
Leonard Bloomfield.
\newblock 1933.
\newblock {\em Language}.
\newblock University of Chicago Press.
\newblock Reprint edition (October 15, 1984).

\bibitem[\protect\citename{Bouchard-C{\^o}t{\'e} \bgroup et al.\egroup
  }2007]{bouchard-EtAl:2007:EMNLP-CoNLL2007}
Alexandre Bouchard-C{\^o}t{\'e}, Percy Liang, Thomas Griffiths, and Dan Klein.
\newblock 2007.
\newblock A probabilistic approach to diachronic phonology.
\newblock In {\em Proceedings of the 2007 Joint Conference on Empirical Methods
  in Natural Language Processing and Computational Natural Language Learning
  (EMNLP-CoNLL)}, pages 887--896, Prague, Czech Republic, June. Association for
  Computational Linguistics.

\bibitem[\protect\citename{Brown \bgroup et al.\egroup
  }1992]{brown1992estimate}
Peter~F. Brown, Vincent J.~Della Pietra, Robert~L. Mercer, Stephen A.~Della
  Pietra, and Jennifer~C. Lai.
\newblock 1992.
\newblock An estimate of an upper bound for the entropy of {E}nglish.
\newblock {\em Computational Linguistics}, 18(1):31--40.

\bibitem[\protect\citename{Carstairs-McCarthy}2010]{carstairs2010evolution}
Andrew Carstairs-McCarthy.
\newblock 2010.
\newblock {\em The Evolution of Morphology}, volume~14.
\newblock Oxford University Press.

\bibitem[\protect\citename{Cotterell \bgroup et al.\egroup
  }2014]{cotterell-peng-eisner:2014:P14-2}
Ryan Cotterell, Nanyun Peng, and Jason Eisner.
\newblock 2014.
\newblock Stochastic contextual edit distance and probabilistic {FST}s.
\newblock In {\em Proceedings of the 52nd Annual Meeting of the Association for
  Computational Linguistics (ACL)}, pages 625--630, Baltimore, Maryland, June.
  Association for Computational Linguistics.

\bibitem[\protect\citename{Cotterell \bgroup et al.\egroup
  }2015]{cotterell-peng-eisner-2015}
Ryan Cotterell, Nanyun Peng, and Jason Eisner.
\newblock 2015.
\newblock Modeling word forms using latent underlying morphs and phonology.
\newblock {\em Transactions of the Association for Computational Linguistics
  (TACL)}, 3:433--447, August.

\bibitem[\protect\citename{Cotterell \bgroup et al.\egroup
  }2017a]{cotterell-conll-sigmorphon2017}
Ryan Cotterell, Christo Kirov, John Sylak-Glassman, G{\'e}raldine Walther,
  Ekaterina Vylomova, Patrick Xia, Manaal Faruqui, Sandra K{\"u}bler, David
  Yarowsky, Jason Eisner, and Mans Hulden.
\newblock 2017a.
\newblock The {CoNLL-SIGMORPHON} 2017 shared task: {U}niversal morphological
  reinflection in 52 languages.
\newblock In {\em Proceedings of the CoNLL-SIGMORPHON 2017 Shared Task:
  Universal Morphological Reinflection}, Vancouver, Canada, August. Association
  for Computational Linguistics.

\bibitem[\protect\citename{Cotterell \bgroup et al.\egroup
  }2017b]{cotterell-sylakglassman-kirov:2017:EACLshort}
Ryan Cotterell, John Sylak-Glassman, and Christo Kirov.
\newblock 2017b.
\newblock Neural graphical models over strings for principal parts
  morphological paradigm completion.
\newblock In {\em Proceedings of the 15th Conference of the European Chapter of
  the Association for Computational Linguistics (EACL)}, pages 759--765,
  Valencia, Spain, April. Association for Computational Linguistics.

\bibitem[\protect\citename{Cotterell \bgroup et al.\egroup
  }2017c]{cotterell-EtAl:2017:EMNLP2017}
Ryan Cotterell, Ekaterina Vylomova, Huda Khayrallah, Christo Kirov, and David
  Yarowsky.
\newblock 2017c.
\newblock Paradigm completion for derivational morphology.
\newblock In {\em Proceedings of the Conference on Empirical Methods in Natural
  Language Processing (EMNLP)}, pages 725--731, Copenhagen, Denmark, September.
  Association for Computational Linguistics.

\bibitem[\protect\citename{Dreyer and Eisner}2009]{dreyer-eisner-2009}
Markus Dreyer and Jason Eisner.
\newblock 2009.
\newblock Graphical models over multiple strings.
\newblock In {\em Proceedings of the Conference on Empirical Methods in Natural
  Language Processing (EMNLP)}, pages 101--110, Singapore, August.

\bibitem[\protect\citename{Dreyer and Eisner}2011]{dreyer-eisner-2011}
Markus Dreyer and Jason Eisner.
\newblock 2011.
\newblock Discovering morphological paradigms from plain text using a
  {D}irichlet process mixture model.
\newblock In {\em Proceedings of the Conference on Empirical Methods in Natural
  Language Processing (EMNLP)}, pages 616--627, Edinburgh, July.
\newblock Supplementary material (9 pages) also available.

\bibitem[\protect\citename{Dreyer \bgroup et al.\egroup
  }2008]{dreyer-smith-eisner:2008:EMNLP}
Markus Dreyer, Jason Smith, and Jason Eisner.
\newblock 2008.
\newblock Latent-variable modeling of string transductions with finite-state
  methods.
\newblock In {\em Proceedings of the 2008 Conference on Empirical Methods in
  Natural Language Processing (EMNLP)}, pages 1080--1089, Honolulu, Hawaii,
  October. Association for Computational Linguistics.

\bibitem[\protect\citename{Edmonds}1967]{edmonds1967optimum}
Jack Edmonds.
\newblock 1967.
\newblock Optimum branchings.
\newblock {\em Journal of Research of the National Bureau of Standards B},
  71(4):233--240.

\bibitem[\protect\citename{Gil}1994]{gil1994}
David Gil.
\newblock 1994.
\newblock The structure of {R}iau {I}ndonesian.
\newblock {\em Nordic Journal of Linguistics}, 17(2):179--200.

\bibitem[\protect\citename{Hare and Elman}1995]{hare1995learning}
Mary Hare and Jeffrey~L. Elman.
\newblock 1995.
\newblock Learning and morphological change.
\newblock {\em Cognition}, 56(1):61--98.

\bibitem[\protect\citename{Hockett}1958]{hockett1958course}
Charles~F. Hockett.
\newblock 1958.
\newblock {\em A Course In Modern Linguistics}.
\newblock The MacMillan Company.

\bibitem[\protect\citename{Kann and Sch\"{u}tze}2016]{kann-schutze:2016}
Katharina Kann and Hinrich Sch\"{u}tze.
\newblock 2016.
\newblock Single-model encoder-decoder with explicit morphological
  representation for reinflection.
\newblock In {\em Proceedings of the 54th Annual Meeting of the Association for
  Computational Linguistics (ACL)}, pages 555--560, Berlin, Germany, August.
  Association for Computational Linguistics.

\bibitem[\protect\citename{Kenstowicz}1994]{kenstowicz1994}
Michael~J. Kenstowicz.
\newblock 1994.
\newblock {\em Phonology in generative grammar}.
\newblock Blackwell Oxford.

\bibitem[\protect\citename{Kibrik}1998]{archi}
Aleksandr~E. Kibrik.
\newblock 1998.
\newblock {A}rchi ({C}aucasian -- {D}aghestanian).
\newblock In {\em The Handbook of Morphology}, pages 455--476. Blackwell
  Oxford.

\bibitem[\protect\citename{Klein \bgroup et al.\egroup }2017]{klein2017opennmt}
Guillaume Klein, Yoon Kim, Yuntian Deng, Jean Senellart, and Alexander~M. Rush.
\newblock 2017.
\newblock {OpenNMT}: {O}pen-source toolkit for neural machine translation.
\newblock {\em arXiv preprint arXiv:1701.02810}.

\bibitem[\protect\citename{McCulloch and Pitts}1943]{mcculloch1943logical}
Warren~S. McCulloch and Walter Pitts.
\newblock 1943.
\newblock A logical calculus of the ideas immanent in nervous activity.
\newblock {\em The Bulletin of Mathematical Biophysics}, 5(4):115--133.

\bibitem[\protect\citename{McWhorter}2001]{mcwhorter2001world}
John McWhorter.
\newblock 2001.
\newblock The world’s simplest grammars are creole grammars.
\newblock {\em Linguistic Typology}, 5(2):125--66.

\bibitem[\protect\citename{Oh}2015]{oh2015linguistic}
Yoon~Mi Oh.
\newblock 2015.
\newblock {\em Linguistic Complexity and Information: Quantitative Approaches}.
\newblock {Ph.D.} thesis, Universit{\'e} de Lyon, France.

\bibitem[\protect\citename{Paz}2003]{paz2003probabilistic}
Azaria Paz.
\newblock 2003.
\newblock {\em Probabilistic Automata}.
\newblock John Wiley and Sons.

\bibitem[\protect\citename{Pellegrino \bgroup et al.\egroup
  }2011]{pellegrino2011across}
Fran{\c{c}}ois Pellegrino, Christophe Coup{\'e}, and Egidio Marsico.
\newblock 2011.
\newblock A cross-language perspective on speech information rate.
\newblock {\em Language}, 87(3):539--558.

\bibitem[\protect\citename{Ralli}1994]{ralli1994feature}
Angela Ralli.
\newblock 1994.
\newblock Feature representations and feature-passing operations in {G}reek
  nominal inflection.
\newblock In {\em Proceedings of the 8th Symposium on English and Greek
  Linguistics}, pages 19--46.

\bibitem[\protect\citename{Ralli}2002]{ralli2002role}
Angela Ralli.
\newblock 2002.
\newblock The role of morphology in gender determination: evidence from modern
  {G}reek.
\newblock {\em Linguistics}, 40(3; ISSU 379):519--552.

\bibitem[\protect\citename{Rissanen and Ristad}1994]{rissanen94}
J.~Rissanen and E.~Ristad.
\newblock 1994.
\newblock Language acquisition in the {MDL} framework.
\newblock In E.~Ristad, editor, {\em Language Computations}. American
  Mathematical Society, Philadelphia.

\bibitem[\protect\citename{Robins}2013]{robins2013}
Robert~Henry Robins.
\newblock 2013.
\newblock {\em A short history of linguistics}.
\newblock Routledge.

\bibitem[\protect\citename{Sagot}2013]{sagot2013comparing}
Beno{\^\i}t Sagot.
\newblock 2013.
\newblock Comparing complexity measures.
\newblock In {\em Computational Approaches to Morphological Complexity}.

\bibitem[\protect\citename{Sapir}1921]{edward1921language}
Edward Sapir.
\newblock 1921.
\newblock Language: An introduction to the study of speech.
\newblock {\em New York: Harcourt, Brace \& Company}.

\bibitem[\protect\citename{Shannon}1948]{Shannon1948}
Claude~E. Shannon.
\newblock 1948.
\newblock A mathematical theory of communication.
\newblock {\em Bell Systems Technical Journal}, 27.

\bibitem[\protect\citename{Siegelmann and Sontag}1991]{siegelmann1991turing}
Hava~T. Siegelmann and Eduardo~D. Sontag.
\newblock 1991.
\newblock Turing computability with neural nets.
\newblock {\em Applied Mathematics Letters}, 4(6):77--80.

\bibitem[\protect\citename{Siegelmann and
  Sontag}1995]{siegelmann1995computational}
Hava~T. Siegelmann and Eduardo~D. Sontag.
\newblock 1995.
\newblock On the computational power of neural nets.
\newblock {\em Journal of Computer and System Sciences}, 50(1):132--150.

\bibitem[\protect\citename{Smith \bgroup et al.\egroup
  }2008]{smith2008introduction}
Kenny Smith, Michael~L. Kalish, Thomas~L. Griffiths, and Stephan Lewandowsky.
\newblock 2008.
\newblock Cultural transmission and the evolution of human behaviour.
\newblock {\em Philosophical Transactions B}.

\bibitem[\protect\citename{Sobel and Li}2013]{sobel2013cognitive}
Carolyn~P. Sobel and Paul Li.
\newblock 2013.
\newblock {\em The Cognitive Sciences: An Interdisciplinary Approach}.
\newblock Sage Publications.

\bibitem[\protect\citename{Spencer}1991]{spencer1991morphological}
Andrew Spencer.
\newblock 1991.
\newblock {\em Morphological Theory: {A}n Introduction to Word Structure in
  Generative Grammar}.
\newblock Wiley-Blackwell.

\bibitem[\protect\citename{Stolz \bgroup et al.\egroup
  }2012]{stolz2012irregularity}
Thomas Stolz, Hitomi Otsuka, Aina Urdze, and Johan van~der Auwera.
\newblock 2012.
\newblock {\em Irregularity in Morphology (and beyond)}, volume~11.
\newblock Walter de Gruyter.

\bibitem[\protect\citename{Sutskever \bgroup et al.\egroup
  }2014]{DBLP:conf/nips/SutskeverVL14}
Ilya Sutskever, Oriol Vinyals, and Quoc~V. Le.
\newblock 2014.
\newblock Sequence to sequence learning with neural networks.
\newblock In {\em Advances in Neural Information Processing Systems (NIPS)},
  pages 3104--3112.

\bibitem[\protect\citename{Sylak-Glassman \bgroup et al.\egroup
  }2015]{sylakglassman-EtAl:2015:ACL-IJCNLP}
John Sylak-Glassman, Christo Kirov, David Yarowsky, and Roger Que.
\newblock 2015.
\newblock A language-independent feature schema for inflectional morphology.
\newblock In {\em Proceedings of the 53rd Annual Meeting of the Association for
  Computational Linguistics and the 7th International Joint Conference on
  Natural Language Processing (ACL)}, pages 674--680, Beijing, China, July.
  Association for Computational Linguistics.

\bibitem[\protect\citename{Sylak-Glassman}2016]{sylak2016composition}
John Sylak-Glassman.
\newblock 2016.
\newblock The composition and use of the universal morphological feature schema
  ({U}nimorph schema).
\newblock Technical report, Johns Hopkins University.

\bibitem[\protect\citename{Zeiler}2012]{zeiler2012adadelta}
Matthew~D. Zeiler.
\newblock 2012.
\newblock {ADADELTA}: {A}n ddaptive learning rate method.
\newblock {\em arXiv preprint arXiv:1212.5701}.

\end{thebibliography}
\bibliographystyle{acl2012}

\end{document}